\pdfoutput=1

\documentclass[11pt]{article}

\usepackage{acl}

\usepackage{listings}

\usepackage{times}
\usepackage{latexsym}
\usepackage{booktabs}
\usepackage[T1]{fontenc}

\usepackage[utf8]{inputenc}

\usepackage{microtype}
\usepackage{hyperref}
\usepackage{url}
\usepackage{times}
\usepackage{amsmath}

\usepackage{cleveref}
\usepackage{latexsym}
\usepackage{float}
\usepackage{amsthm}
\usepackage{tcolorbox}
\usepackage{microtype}
\usepackage{amsmath, amssymb}
\usepackage{booktabs}
\usepackage{wrapfig}
\usepackage{graphicx}
\usepackage{multirow}
\usepackage{inconsolata}
\usepackage{graphicx}
\usepackage{tikz}
\usepackage{pgfplots}
\pgfplotsset{compat=newest}
\usepackage{graphicx}

\title{Enhancing Fine-Grained Image Classifications via Cascaded Vision Language Models}

\author{Canshi Wei \\ Tencent \\ \texttt{canshiwei@gmail.com}}

\begin{document}
\maketitle
\begin{abstract}
Fine-grained image classification, particularly in zero/few-shot scenarios, presents a significant challenge for vision-language models (VLMs), such as CLIP. These models often struggle with the nuanced task of distinguishing between semantically similar classes due to limitations in their pre-trained recipe, which lacks supervision signals for fine-grained categorization. This paper introduces CascadeVLM, an innovative framework that overcomes the constraints of previous CLIP-based methods by effectively leveraging the granular knowledge encapsulated within large vision-language models (LVLMs).
Experiments across various fine-grained image datasets demonstrate that CascadeVLM significantly outperforms existing models, specifically on the Stanford Cars dataset, achieving an impressive 85.6\% zero-shot accuracy.
Performance gain analysis validates that LVLMs produce more accurate predictions for challenging images that CLIPs are uncertain about, bringing the overall accuracy boost. Our framework sheds light on a holistic integration of VLMs and LVLMs for effective and efficient fine-grained image classification. 
\end{abstract}



\section{Introduction}
\begin{figure}[t!]
\centering
\includegraphics[width=0.9\linewidth]{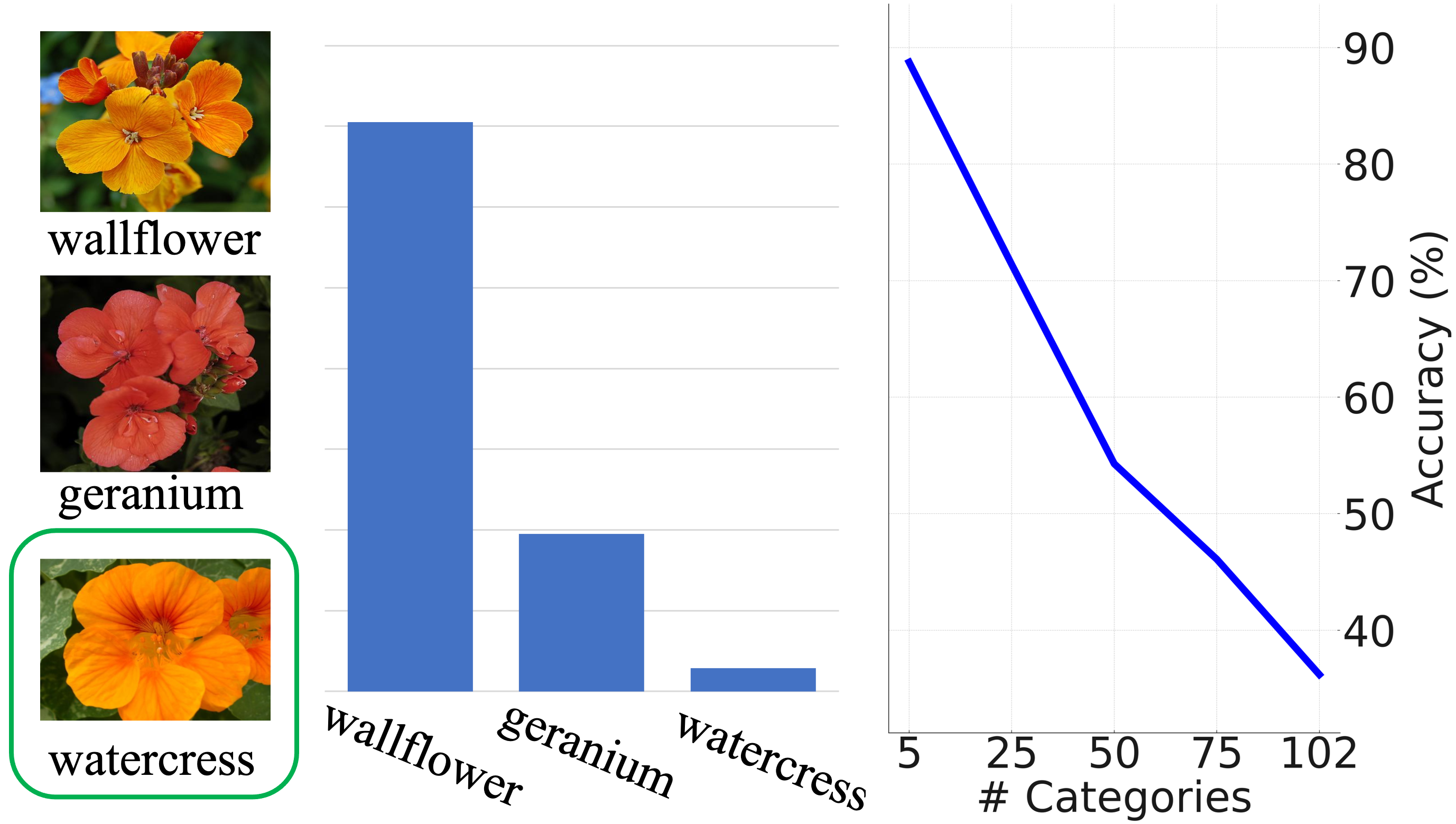}
\caption{Illustration of model performance: CLIP's misclassification of watercress (left) and the inverse relationship between LVLM accuracy and the number of categories (right).}
\label{fig:flowerbase}
\end{figure}

The vision-language model (VLM) landscape is evolving, highlighted by models like CLIP~\citep{radford2021clip} and its significant zero/few-shot classification capabilities~\citep{zhou2022learning}. Despite these advances, fine-grained image classification remains challenging, particularly in distinguishing closely related sub-classes~\citep{ren-etal-2023-delving}, such as the \textit{watercress} and the \textit{wallflower} in Figure~\ref{fig:flowerbase}, due to the subtle visual differences between classes.
To enhance the zero/few-shot fine-grained classification performance of CLIP, recent methods have concentrated on advanced prompt engineering~\citep{DBLP:journals/corr/abs-2109-01134,zhou2022conditional} and the enhancement of pre-training supervision~\citep{li2023scaling,singh2023effectiveness}. Additionally, to enrich CLIP's prompt context, \citet{menon2022visual} introduced a technique whereby GPT3 generates detailed class descriptions for enhancing CLIP models. 
However, 
this method cannot handle visually similar classes as the generated descriptions tend to be nearly identical, offering limited benefits for fine-grained classification tasks.

In this paper, we instead resort to large vision language models (LVLMs), to better capitalize on their extensive world knowledge.
A straightforward way is to directly concatenate all the candidate classes as a prompt for LVLMs. However, the limited capacity for long-context modeling~\citep{zhao2023survey} poses challenges, particularly evident when LVLMs grapple with a large candidate image class set. As shown in  Figure~\ref{fig:flowerbase}, an advanced open-sourced LVLM, Qwen-VL~\citep{Qwen-VL}, still suffers from a dramatic accuracy decline when the candidate categories increased from 5 to 102. 
To tackle this, we employ CLIP models for locating potential candidate classes to ease the burden for LVLMs. 
Due to the contrastive pre-training objective, CLIP is capable of finding a set of possible classes, evidenced by the relatively higher top-K prediction accuracy. 
For example, in the Flowers102 dataset, CLIP(ViT-B/32) gives a 68.7\% Top-1 accuracy, yet the Top-10 accuracy is significantly boosted to 89.9\%. This characteristic validates our motivation to holistically integrate CLIP and LVLM. 

Motivated by our previous exploration, in this paper, we introduce CascadeVLM, which integrates the complementary capabilities of CLIP-like models and LVLMs to perform fine-grained image classification. The key idea is to leverage the CLIP-like models as a class filter for LVLMs to fulfill the LVLM potentials.
Besides, the results can be further enhanced by leveraging the in-context learning~\citep{icl_survey} of LVLMs to perform few-shot learning. Moreover, the overall inference efficiency can be improved by adopting an entropy threshold as a heuristic mechanism to evaluate the necessity of deploying LVLMs, achieving a dynamic early exiting~\citep{xin2020deebert,li2021cascadebert}.
We conducted zero/few-shot experiments across various fine-grained image datasets, and the evaluation results demonstrate that CascadeVLM surpasses other individual models in performance. For instance, CascadeVLM achieved an 85.6\% accuracy rate on the Stanford Cars dataset. Additionally, we utilized two advanced pretrained CLIP-like models as backbones for the iNaturalist and SUN397 datasets, achieving superior results compared to the individual backbone models.
Furthermore, our analysis meticulously dissects CascadeVLM's performance improvements, spotlighting the model's sophisticated handling of entropy thresholds to balance computational demands with accuracy. This dual-focused investigation not only clarifies the operational dynamics of CascadeVLM but also its adeptness in integrating with CLIP-like architectures and LVLM for refined final predictions, underlining the model's comprehensive adaptability and strategic efficiency.

In summary, the contributions of this paper could be summarized as follows:
(1) We propose the CascadeVLM framework which effectively combines CLIP-like VLMs and LVLMs for zero/few-shot fine-grained image classification.
(2) Experiment on six relevant datasets reveals that CascadeVLM achieves performance on par with advanced zero/few-shot methods. Our analysis provides insights into the holistic integration of VLMs and LVLMs

\begin{figure*}[t!]
\centering
\includegraphics[width=\textwidth]{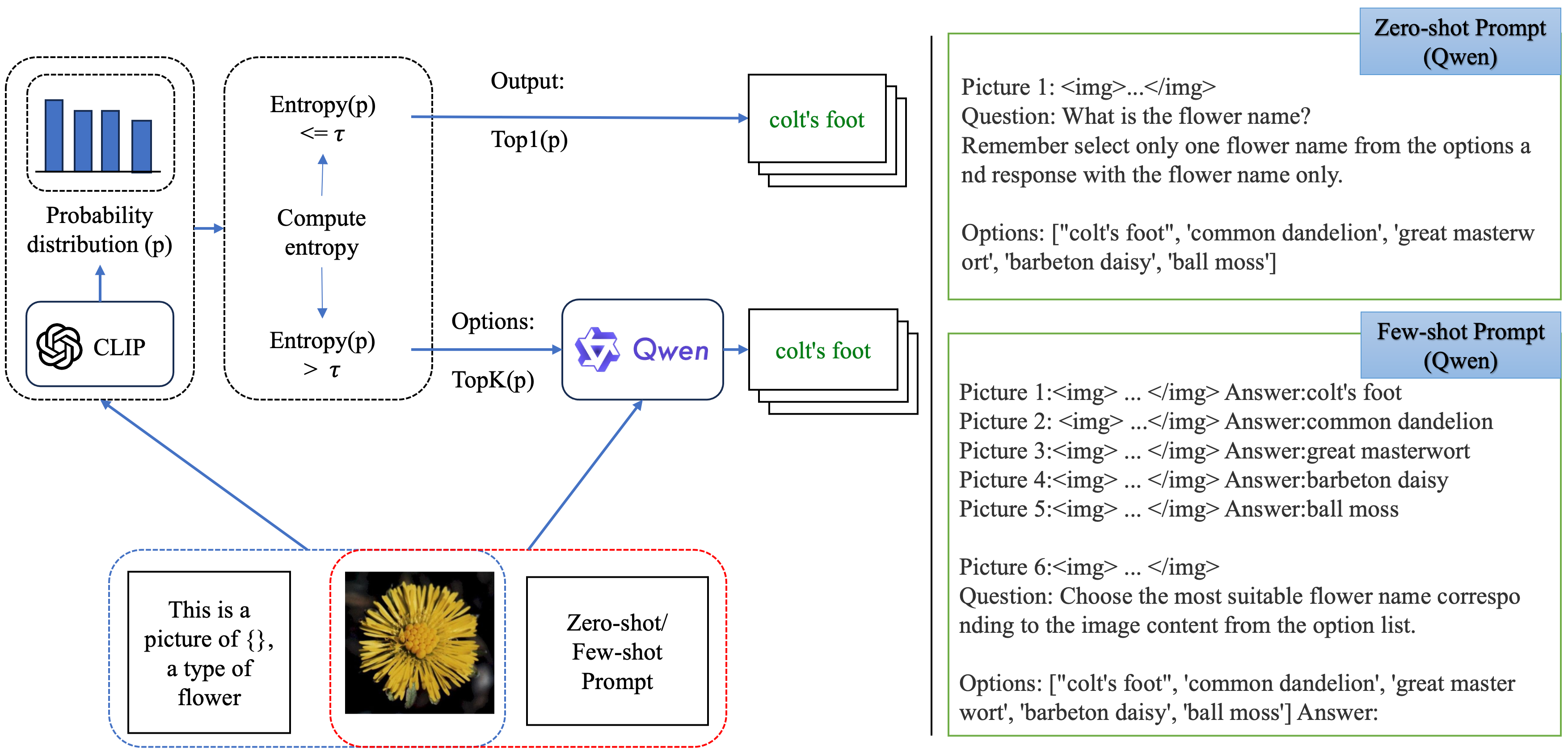}
\caption{CascadeVLM commences with CLIP for initial image analysis and probabilistic categorization, integrating an entropy threshold, $\tau$, to balance efficiency and accuracy, culminating in LVLM's adaptive classification.}
\label{fig:cascadeVLM}
\end{figure*}

\section{Methodology}

In this section, we delineate the methodology underpinning our CascadeVLM framework, which is structured into two steps. (1) The first step involves candidate selection facilitated by the CLIP model. (2) The second step encompasses the application of zero-shot or few-shot prediction techniques using large vision-language models (LVLMs). For zero-shot prediction, LVLMs directly engage in classification based on these pre-selected candidates. For few-shot situations, we use the in-context learning strategy to augment the semantic context. 

\subsection{CLIP-based Candidate Selection}

As a pivotal component of our CascadeVLM framework, CLIP's operational mechanism~\cite{radford2021clip} allows it to effectively discern potential correct classes, making it an ideal choice for the initial phase of candidate filtering from an extensive array of class labels.

Specifically, in our approach, the function $f_{\text{CLIP}}(x, c_i)$ denotes the score outputted by the CLIP model for a specific category $c_i$ when given an image $x$. CLIP’s core functionality lies in its ability to align image and text representations within a unified embedding space, and gives scores for each pairs. Upon acquiring raw scores from CLIP for each category in the label set $C$, we employ a softmax function to transform these scores into a probability distribution, as delineated by: 
\begin{equation}
    P(c_i \mid x) = \frac{\exp(f_{\text{CLIP}}(x, c_i))}{\sum_{c_j \in C} \exp(f_{\text{CLIP}}(x, c_j))},
    \label{eq1}
\end{equation}
The resulting probabilities could reflect the relative confidence of the CLIP model in associating the given image with each category within the context of the entire set $C$. 

Based on the probability computation, denoted as $P(c_i \mid x)$, specified in Equation~\ref{eq1}, we extract the top-$k$ categories from $C$ to ensure that they are sorted in descending order of probability. This selection and sorting process, crucial for the framework's efficacy, is denoted as a function $s_{\text(topk)}$. Not only does this step condense the pool of candidate classes, but it also addresses the sensitivity of LVLMs to the sequence in which these categories are presented. Our empirical results~\ref{subsec:result_zero_shot} affirm that simple sorting based on probability significantly bolsters the predictive precision of LVLMs. The generalized representation of this procedure is as follows:
\begin{equation}
    C^* = \{c'_1, c'_2, \ldots, c'_k\} = s_{\text{topk}}(P(c_i \mid x), C),
\end{equation}
where $C^*$ encapsulates the optimally sorted candidates, with $c'_1$, $c'_2$ through to $c'_k$ representing the elements in descending order of their computed probabilities.



\subsection{LVLMs Prediction with Candidate Set}


In this section, we seek to leverage Large Vision-Language Models (LVLMs) for the final classification within our CascadeVLM framework. Capitalizing on a subset of candidates pre-selected by CLIP, LVLMs overcome the challenge of extensive context and improve prediction accuracy through adaptable zero-shot and few-shot learning strategies, tailored to the data-sparse environments.

\paragraph{Zero-Shot Prediction}

Zero-shot learning~\cite{NIPS2013_2d6cc4b2} enables models to predict unseen classes without specific training examples, leveraging pre-existing knowledge from broader contexts or related tasks. This method is particularly beneficial in data-scarce scenarios, where it effectively infers new categories despite limited training data. \looseness=-1

In the context of our CascadeVLM framework, zero-shot prediction is executed after identifying the top-$k$ candidate classes using CLIP. The LVLM then selects one candidate $c*$, as the final prediction. Here, we generalize the process of LVLM prediction as function $f_{\text(LVLM)}$, given the input image $x$ and the top-$k$ candidate set $C^*$:

\begin{equation}
    c^* = f_{\text{LVLM}}(x, C^*)
\end{equation}


The zero-shot prediction phase in our CascadeVLM framework highlights LVLMs' proficiency in utilizing pre-trained knowledge for unseen data while adeptly managing contextual complexities. Focusing on a select set of candidates, the proposed methods coude effectively address the intricacies of fine-grained classification, ensuring precise and dependable outcomes even without class-specific examples.

\paragraph{Few-Shot Prediction}
In the Few-Shot Prediction phase of our CascadeVLM framework, we capitalizes on LVLMs' in-context learning~\cite{NEURIPS2020_1457c0d6} ability, where additional relevant samples significantly enhance performance, allowing LVLMs to deepen their understanding and improve predictive accuracy.

In the integration of few-shot learning within our cascade framework, we undertake a two-step process for candidate categories set $C^*$:

\emph{Step 1: Context Generation:}
In this initial phase, for each category $c'_i$ in $C^* $, we randomly select an example image $x_{c'_i} $ from the training dataset, and design a prompt to contextualize the input image $x$ for the LVLMs. Here, each candidate class $c'_i$ and its corresponding example image $x_{c'_i}$ are integrated with the prompt template, creating a contextual framework for the LVLMs. We denote this assemblage of prompts and images form the contextual basis as $E$ in the subsequent step of our methodology. For instance, within the context of the GPT4-V scenario, the contextual basis denoted as $E$ is formulated in Table~\ref{tab:few-shot prompt}.

\begin{table}[h]
    \centering
    \begin{tcolorbox}
\texttt{<IMG:} \; $x_{c'_1}$\texttt{>} \\
\texttt{Question: What is the class of the image?}  \texttt{Answer:} ${c'_1}$ \\
\texttt{<IMG:} \; $x_{c'_2}$\texttt{>} \\
\texttt{Question: What is the class of the image?}  \texttt{Answer:} ${c'_2}$ \\
\vdots
    \end{tcolorbox}
    \caption{Few-shot prompt used in our experiments.}
    \label{tab:few-shot prompt}
\end{table}

\emph{Step 2 - Prediction with Contextual Information:}
In this step, the context $E$ is integrated with the input image $x$ and fed into the LVLMs. This integration enables the LVLMs to utilize the rich contextual information embedded in $E$ to enhance and refine its predictive process for the image $x$. Consequently, the final classification outcome, denoted as $c^*$, emerges from this enriched inferential framework. The process can be mathematically represented as:
\begin{equation}
    c^* = f_{\text{LVLM}}(x, C^*, E)
\end{equation}
where \( f_{\text{LVLM}} \) represents the LVLM prediction based on provided image $x$, the top-$k$ candidate set $C^*$ and the context set $E$.



\subsection{Speed-up via Adaptive Entropy Threshold}

In our CascadeVLM framework, we introduce an adaptive entropy-based approach aimed at enhancing inference speed, reducing the computational load on LVLMs, and accelerating overall throughput. The entropy $H(x)$ of the probability distribution, a measure of uncertainty or predictability within the distribution, is calculated as follows:

\begin{equation}
    H(x) = -\sum_{c_i \in C} P(c_i \mid x) \log P(c_i \mid x)
\end{equation}

This computation serves as a critical decision point in our methodology. If the calculated entropy $H(x)$ falls below a predefined threshold, it signifies a high confidence level in the top-1 category as determined by CLIP. In such cases, we expedite the process by directly outputting this top-1 category, thereby bypassing the need for further LVLM processing. Conversely, if $H(x)$ exceeds the threshold, indicating a lower level of confidence and greater uncertainty, we proceed to the subsequent steps involving LVLMs for refined classification. This adaptive mechanism effectively balances speed and accuracy, streamlining the framework while ensuring reliable classification outcomes.


\section{Experiments}

In this section, we rigorously evaluate the performance of our CascadeVLM framework across diverse benchmarks. Initially, we detail the experimental setup in Section~\ref{subsec:exp_setup}, followed by an in-depth analysis of the framework's efficacy in zero-shot learning scenarios in Section~\ref{subsec:result_zero_shot}, and subsequently in few-shot learning contexts in Section~\ref{subsec:result_few_shot}. 


\begin{table}[t!]
    \centering
    \scalebox{0.9}{ 
    \begin{tabular}{l|cc}
    \toprule 
    Dataset & \# of Class & \# of Test  \\
    \midrule 
    Flowers102 & 102 & 818 \\
    StanfordCars & 196 & 8,041 \\
    FGVC Aircraft & 100 & 3,333 \\
    BirdSnap & 500 & 2,444 \\
    SUN397 & 397 & 19,850 \\
    iNat18 (iNaturalist 2018) & 8,142 & 24,426 \\
    \bottomrule
    \end{tabular}
    } 
    \caption{Statistics of the evaluated fine-grained image classification benchmarks.}
    \label{tab:datasets summary}
\end{table}

\begin{table*}[t!]
    \centering
    \small 
    \begin{tabular}{l|cccc|c}
    \toprule 
    Model & Flower102 & StanfordCars & FGVC Aricraft & BirdSnap & Avg.\\
    \midrule
    \textcolor{gray}{Supervised} & \textcolor{gray}{99.8} \shortcite{DBLP:journals/corr/abs-2104-05704} & \textcolor{gray}{96.3}~\shortcite{DBLP:journals/corr/abs-2104-10972} & \textcolor{gray}{95.4}~\shortcite{Bera_2022} & \textcolor{gray}{90.1}~\shortcite{DBLP:journals/corr/abs-2010-01412} & - \\ 
    \midrule 
    Qwen (full classes) & 37.5 & 22.4 & 8.4 & 2.3 & 17.6 \\
    
    \midrule
    CLIP ViT-B/32 & 68.7 & 59.6 & 19.1 & 51.7 & 49.8 \\ 
    CLIP ViT-B/32 Cascade (Qwen, full classes) & 72.7 & 74.3 & 22.7 & 20.3 & 47.5 \\
    CLIP ViT-B/32 Cascade (Qwen, top $k$) & \textbf{74.2} & \textbf{79.2} & \textbf{27.1} & \textbf{56.7} & \textbf{59.3} \\ 
    \midrule
    CoOp ViT-B/16~\citep{DBLP:journals/corr/abs-2109-01134} & 68.7 & 64.5 & 18.5 & - & - \\
    CoCoOp ViT-B/16~\citep{zhou2022conditional} & 71.9 & 65.3& 22.9 & - & - \\
    POMP ViT-B/16~\citep{ren2023prompt} & 72.4 & 66.8 & 25.6 & - & - \\
    CLIP ViT-B/16 & 73.0 & 64.4 & 24.5 & 52.5 & 53.6 \\ 
    CLIP ViT-B/16 Cascade (Qwen, full classes) & 70.5 & 74.1 & 26.2 & 20.2 & 47.8 \\ 
    CLIP ViT-B/16 Cascade (Qwen, top $k$) & \textbf{73.3} & \textbf{79.1} & \textbf{30.7} & \textbf{56.6} & \textbf{60.0} \\
    \midrule
    FLIP ViT-L/14~\citep{li2023scaling} & 75.0 & \textbf{90.7} & 29.1 & \textbf{63.0} & 64.5 \\
    CLIP ViT-L/14 & \textbf{81.3} & 76.2 & 30.9 & 62.2 & 62.7 \\ 
    CLIP ViT-L/14 Cascade (Qwen, full classes) & 75.8 & 78.9 & 30.1 & 21.0 & 51.5 \\
    CLIP ViT-L/14 Cascade (Qwen, top $k$) & 78.2 & 85.6 & \textbf{37.0} & \textbf{63.0} & \textbf{66.0} \\
    \bottomrule 
    \end{tabular}
    \caption{Zero-shot results comparison with different CLIP models as the backbone. The $k$ is selected based on the validation set. Our CascadeVLM achieves the best overall performance on four benchmarks. The best scores are shown in \textbf{bold}.}
    \label{tab: Qwen cascade}
\end{table*}


\subsection{Experimental Settings}
\label{subsec:exp_setup}

\paragraph{Models} 


For experimental evaluation, we employed various CLIP models in combination with specific Large Vision-Language Models (LVLMs). The experiments utilized one of the CLIP variants CLIP ViT-B/32, ViT-B/16, or ViT-L/14 alongside either Qwen-VL-Chat~\citep{Qwen-VL} or GPT-4V~\citep{gpt4v} as the LVLM. Additionally, we explore the framework's adaptability by integrating it with two robust CLIP-like models, MAWS-CLIP~\citep{singh2023effectiveness} and CLIP (ViT-G/14) pre-trained with OpenCLIP~\citep{ilharco_gabriel_2021_5143773} using the Laion2B~\citep{schuhmann2022laionb} dataset, as backbones cascading with GPT-4V. 


\paragraph{Datasets}

We utilize a collection of datasets each offering unique characteristics and significance for fine-grained image classification, as summarized in Table~\ref{tab:datasets summary}. These datasets include Flowers102~\citep{flower_dataset}, StanfordCars~\citep{stanfordcars_dataset}, FGVC Aircraft~\citep{fgvc_aircraft_dataset}, BirdSnap~\citep{birds525_dataset}, SUN39~\citep{Xiao:2010}, and iNaturalist(2018)~\citep{vanhorn2018inaturalist}, collectively encompassing a wide range of categories. 


\paragraph{Baselines}


In our experimental evaluation, we benchmark the performance of various CLIP variants both individually and within our cascade framework. This comparative analysis allows us to demonstrate the enhancement in classification accuracy achieved by integrating these models into the CascadeVLM framework. Moreover, we extend our comparison to include diverse methodologies, providing a comprehensive evaluation of our proposed approach.

\begin{table}[t!]
    \centering
    \small 
    \begin{tabular}{l|cc}
    \toprule 
    Model & iNat18 & SUN397 \\
    \midrule
    MAWS-CLIP & 20  & 71.0 \\ 
    MAWS-CLIP Cascade & \textbf{26.8} & \textbf{75.4} \\ 
    \midrule
    OpenCLIP (ViT-G/14) & 6.6 & 74.4  \\ 
    OpenCLIP (ViT-G/14) Cascade & \textbf{11.8} & \textbf{77.6}  \\ 
    \bottomrule
    \end{tabular}
    \caption{Zero-shot prediction result comparison with MAWS and OpenCLIP ViT-G/14 as backbone, cascading GPT-4V.}
    \label{tab:maws_openclip_gpt4v}
\end{table}

\begin{table*}[t!]
    \centering
    \small 
    \begin{tabular}{l|cccc|c}
    \toprule
    Model & Flower102 & StanfordCars & FGVC Aricraft & BirdSnap & Avg. \\
    \midrule 
    CLIP(ViT-L/14) & 82.0 & 75.0 & 30.0 & 60.5 & 61.9 \\ 
    GPT4-V (full classes) & 67.5 & 74.0 & 61.5 & 46.0 & 62.3 \\
    CLIP(ViT-L/14) + GPT4-V (k=full classes) & 82.0 & 82.5 & \textbf{64.5} & 55.5 & 71.1\\
    CLIP(ViT-L/14) + GPT4-V (k=5) & 86.5 & 85.5 & 56.0 & 62.0 & 72.5 \\ 
    CLIP(ViT-L/14) + GPT4-V (k=5) + 1-shot & \textbf{94.5} & \textbf{88.5} & 63.0 & \textbf{72.5} & \textbf{79.7} \\
    \bottomrule
    \end{tabular}
    \caption{Few-shot learning results with GPT-4V as the LVLM. GPT-4V can better utilize the in-context demonstrations to achieve superior results for fine-grained classification. The result of CasecadeVLM is superior overall datasets.}
    \label{tab: gpt4v cascade}
\end{table*}

\subsection{Zero-shot Learning Results}
\label{subsec:result_zero_shot}

Table \ref{tab: Qwen cascade} showcases the zero-shot prediction capabilities of CascadeVLM, demonstrating its superior performance across several benchmarks. Remarkably, without necessitating any training, CascadeVLM outperforms established methods such as CoOp, CoCoOp, and POMP. Moreover, it is comparable to and often outperforms the more strongly supervised method FLIP, which leverages fine-grained supervision for training. Given the flexibility of CascadeVLM, there is potential for further accuracy enhancements by employing FLIP as the underlying architecture. Consequently, we propose an experimental framework that integrates two advanced pre trained CLIP-like models as the backbone to explore this potential.



Table \ref{tab:maws_openclip_gpt4v} presents the results of experiments that applied two advanced CLIP-like models, MAWS CLIP and OpenCLIP (ViT-G/14), cascading to GPT-4V to challenging datasets: iNaturalist and SUN397, with random samples of 500 from each due to budget and rate limit constraints. The imbalanced nature of iNaturalist typically poses challenges for CLIP models. However, the findings reveal significant performance enhancements when utilizing a cascade framework. Specifically, MAWS CLIP achieved improvements of 6.8\% on iNaturalist and 3.6\% on SUN397. Similarly, OpenCLIP's performance increased by 5.2\% on iNaturalist and 3.2\% on SUN397. These improvements highlight the cascade framework's adaptability and effectiveness in enhancing the capabilities of CLIP-like models.

\subsection{Few-shot Learning Results}
\label{subsec:result_few_shot}

In our initial explorations, we assessed Qwen-VL's capacity for few-shot learning within fine-grained image classification domains. However, it became apparent that Qwen-VL struggled to optimally utilize in-context demonstrations and instructions in this setting. Consequently, we turned our focus to GPT-4V, anticipating its better alignment with our framework's requirements.

Our experiments with GPT-4V were limited to a random subset of 200 samples per dataset due to budget and rate limit constraints. Here we utilize top-k (k=5) for few-shot experiments and yield even more pronounced improvements in predictive accuracy. For instance, with few-shot learning applied, the Flower102 dataset achieved an impressive 94.5\% accuracy, while the StanfordCars dataset attained 88.5\%. These results not only reaffirm the effectiveness of our cascade framework but also highlight its adaptability and efficiency in leveraging few-shot learning for fine-grained classification tasks.



\section{Analysis}


In this section, we explore various aspects of CascadeVLM, highlighting the underlying reasons for its enhanced performance, the trade-off of the entropy threshold, and more. More investigations can be found in the Appendix~\ref{sec:appendix_analysis}.

\begin{figure}[t!]
\centering
\includegraphics[width=\linewidth]{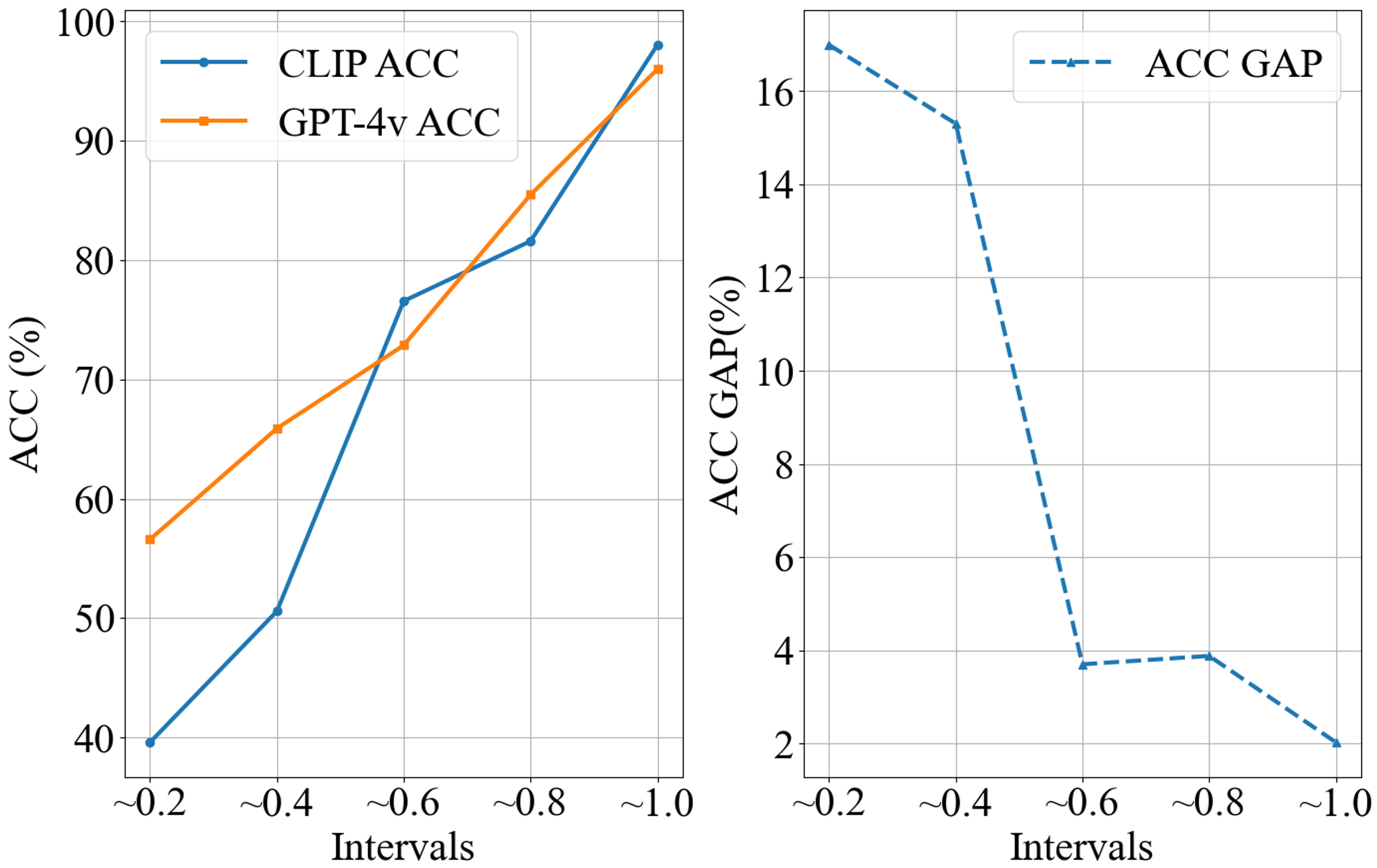}
\caption{Comparative Analysis of ACC performance between CLIP and GPT-4V across different intervals of classification certainty. The left graph shows the ACC of both models across varying levels of margin. The right graph presents the ACC gap between the two models.}
\label{fig:perf_analysis}
\end{figure}

\subsection{Performance Gain Analysis}
\label{subsec:perf_analysis}

This analysis seeks to demonstrate why cascading CLIP with an LVLM model leads to improved accuracy in classification tasks. Using the Flowers102 dataset as a case study, we assess the performance of CLIP and the enhancement brought by LVLM. The margin~\citep{settles2009active}, i.e., the difference between the top1 and top2 probability scores from CLIP, serves as an indicator of the model's certainty about its prediction, where smaller margins suggest greater ambiguity in the image classification.

We divide the range of margins into five intervals, from 0 to 1, to analyze the effects systematically. The data reveals that GPT-4V, representing LVLM, significantly outperforms CLIP with margins less than 0.4, where CLIP experiences confusion. 
This is evident in the consistently high ACC for GPT-4V in these instances. When the margin exceeds 0.6, the ACC for CLIP improves, indicating that the model is more confident and accurate in its predictions, thus reducing the gap in performance between CLIP and LVLM. The accompanying Figure~\ref{fig:perf_analysis} illustrates this trend, with the ACC gap decreasing sharply as the margin increases. This pattern suggests that while LVLM provides a significant advantage in cases of high ambiguity, the benefit tapers off as CLIP's confidence in its classifications rises.

\begin{figure}[t!]
\centering
\includegraphics[width=\linewidth]{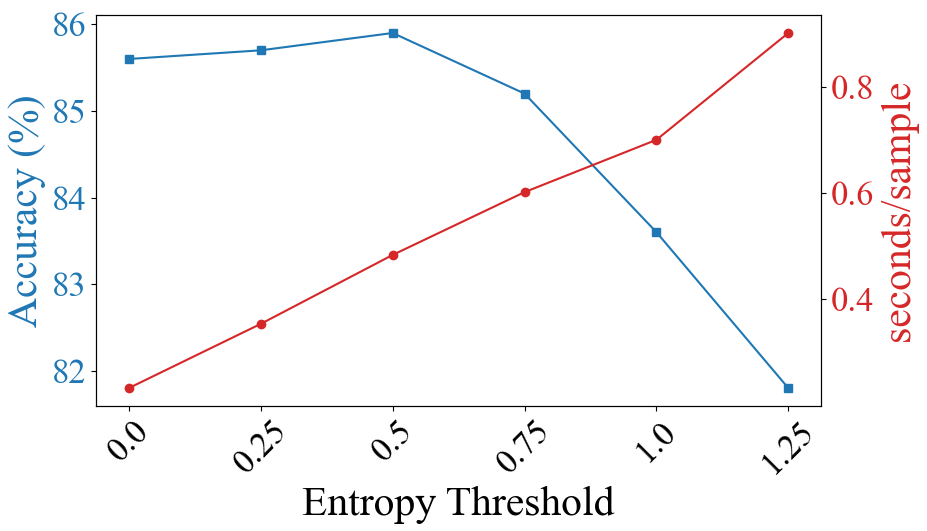}
\caption{Performance variation in the StanfordCars dataset with varying entropy thresholds using CLIP-ViT-L/14 for cascading, set at top-k=10. An increase in entropy threshold results in decreased inference speed and reduced accuracy.}
\label{fig:entropy_threshold}
\end{figure}

\subsection{Sensitivity of Option Orders}
As delineated in Tables \ref{tab: Qwen cascade} and \ref{tab: gpt4v cascade}, we find, surprisingly, that the arrangement of options provided by CLIP plays a pivotal role in the efficacy of Language-Vision Language Models (LVLMs). While it may seem a minor detail to supply LVLMs with the entire class set from CLIP, this procedure is significantly impactful. For example, as shown in Table \ref{tab: Qwen cascade}, presenting all classes in a random order to Qwen results in an average accuracy of only 17.6\%. Conversely, when the classes are organized according to the probabilities assigned by CLIP, there is a notable enhancement in performance. Specifically, in the case where CLIP(ViT-L/14) cascades with Qwen, offering a fully ordered class set, there is a substantial accuracy increase of 33.9\% across various datasets. This improvement is consistently observed with GPT-4V as well, where organizing the classes based on CLIP's guidance boosts accuracy by 8.8\% across the datasets.

\subsection{Inference Efficiency Analysis}
\label{subsec:threshold_analysis}

This subsection critically evaluates the efficacy of implementing an entropy threshold within the CascadeVLM framework. Functioning as a heuristic determinant, this threshold crucially dictates the juncture at which processing shifts from CLIP's initial evaluation to the computationally demanding LVLM analysis. This strategic integration plays a pivotal role in augmenting inference speed, adeptly balancing expeditious processing with the need for in-depth LVLM processing. Our experiments, conducted in a 1GPU (V100) environment, are illustrated in Figure~\ref{fig:entropy_threshold}. Results indicate a direct correlation between an increase in entropy threshold and heightened inference speed, albeit at the cost of reduced accuracy.

\begin{figure}[t!]
\centering
\includegraphics[width=\linewidth]{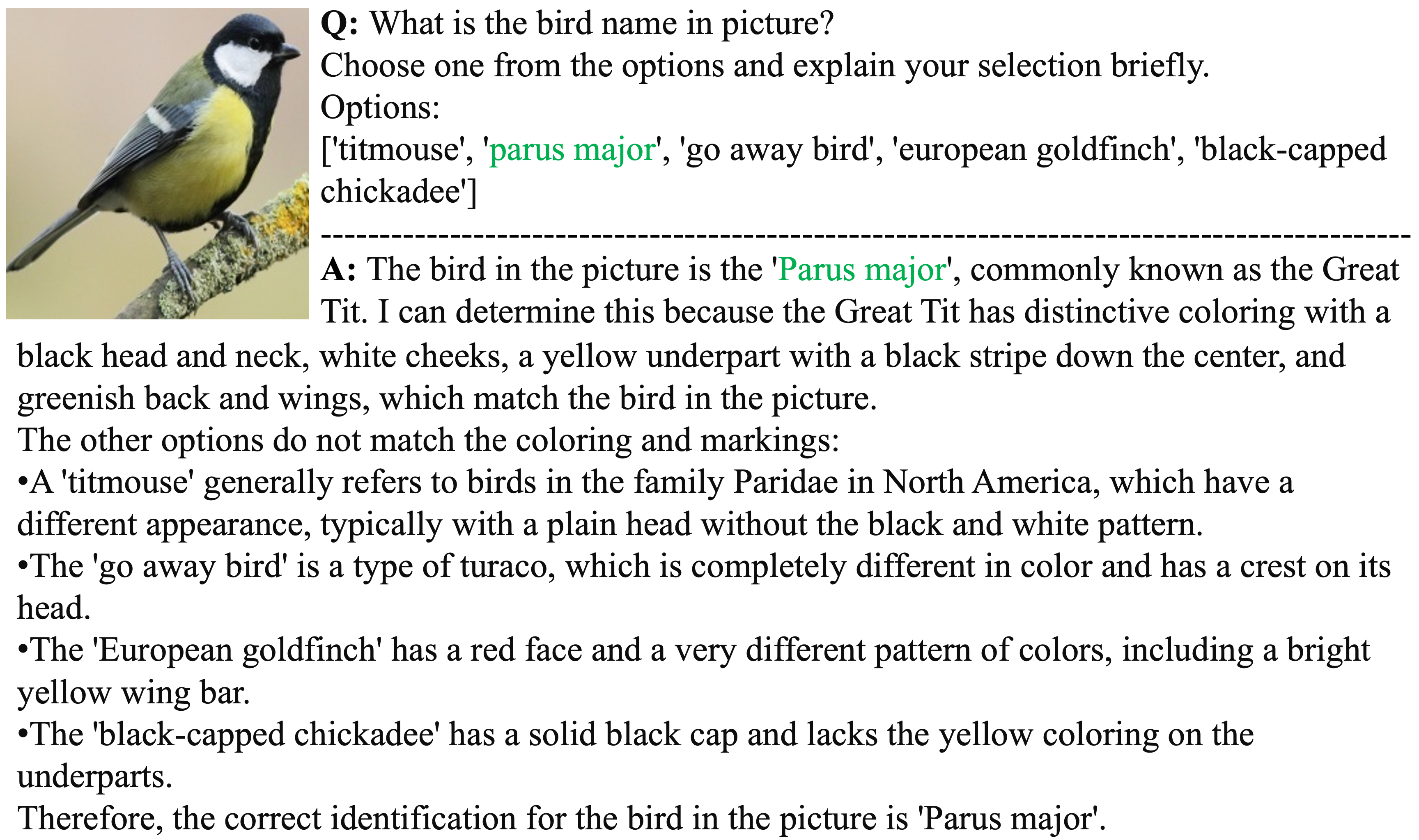}
\caption{CLIP's prediction is incorrect, while LVLM corrects the answer and makes the reasoning.}
\label{fig:reasoning}
\end{figure}

\subsection{Towards Explainable Predictions}
Understanding the rationale behind the predictions in image classification provides valuable insights for a robust decision~\citep{chen2019looks}.
Employing the cascade framework, we design an explanation prompt that inquires why the LVLM prioritizes one option over others. 
For instance, as depicted in Figure~\ref{fig:reasoning}, CLIP initially predicts an incorrect answer. 
The LVLM not only corrects this but also explains its prediction, allowing us to dissect the LVLM’s reasoning ability. 
This enables us to differentiate between the terms \emph{titmouse} which is CLIP’s prediction, and \emph{parus major}, which is the LVLM's corrected choice.
This case highlights the great potential of our CascadeVLM beyond producing accurate classification results.

\subsection{Case Study}
\label{subsec:case_study}

Our case study analysis presents an examination of three distinct scenarios encountered during experimentation with CLIP ViT-L/14 and Qwen as the LVLM in a $k=5$ setting.

Case 1 illustrates a scenario where CLIP's top-1 prediction is incorrect; however, the ground truth is present within its top-5 predictions. Leveraging the LVLM's discernment, the accurate class—green-winged dove—is selected.

Case 2 depicts a situation where, despite CLIP's inclusion of the correct answer—striped owl—in its top-5 predictions, the LVLM fails to identify it correctly. This instance highlights potential areas for refinement within the LVLM's decision-making process.

Case 3 demonstrates a complete misalignment where both CLIP and LVLM fail to recognize the correct class within the top-5 predictions, leading to a compounded error in the final outcome.

These cases underscore the nuanced complexities of fine-grained image classification and reaffirm the necessity for integrated approaches like CascadeVLM to capitalize on the strengths of both CLIP and LVLMs. They also provide valuable insights into the decision-making dynamics of the models, offering pathways for future enhancements.

\begin{figure}[t!]
\centering
\includegraphics[width=\linewidth]{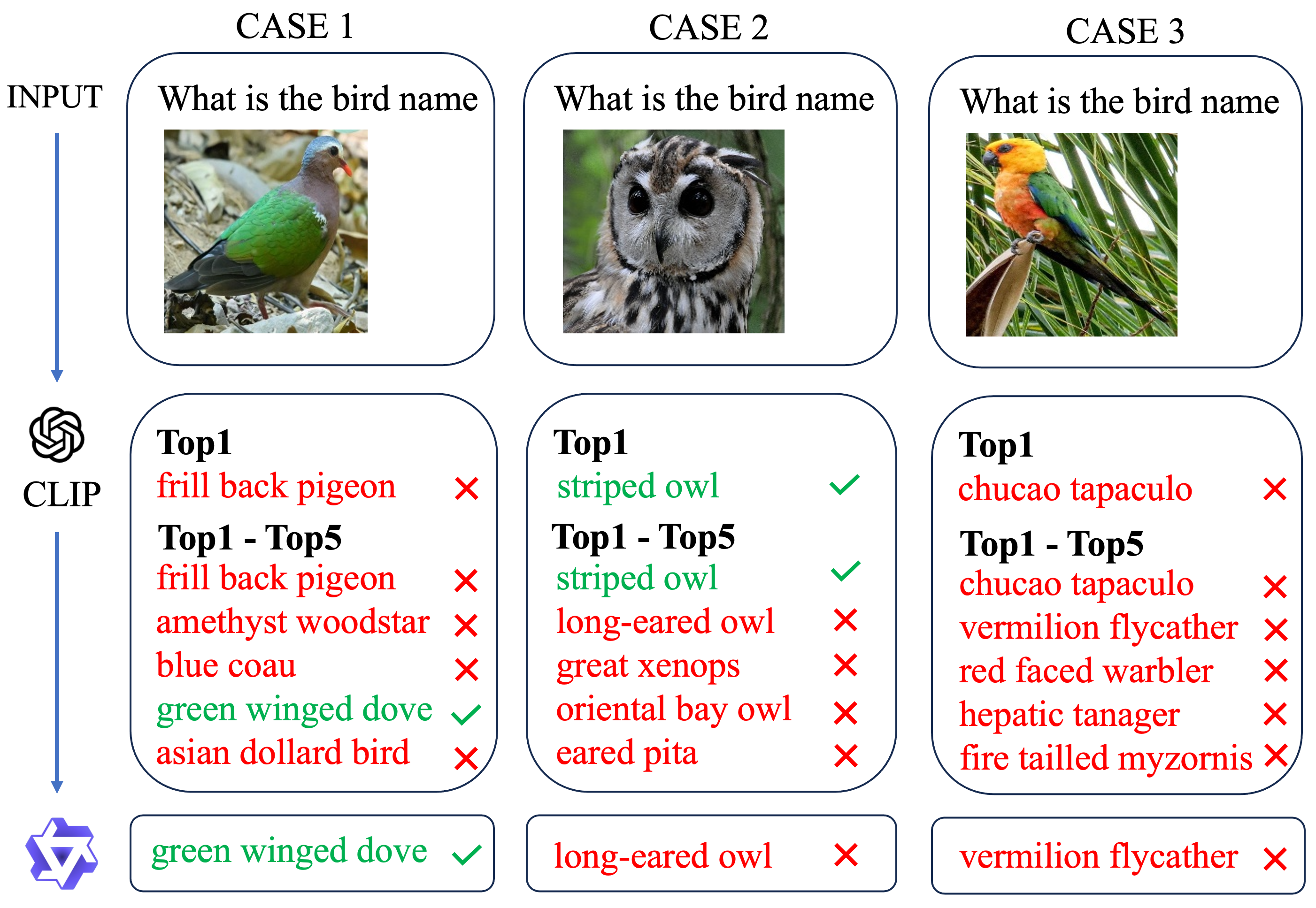}
\caption{Three case studies demonstrating the cascade process from CLIP predictions to LVLM refinement for bird species classification.}
\label{fig: caseStudy}
\end{figure}

\section{Related Work}

\paragraph{Vision Language Models}
Building vision language models (VLMs) for understanding the multi-modal world has been an active research area. Pilot studies leverage pre-training concepts from NLP~\citep{devlin2019bert}, learning shared representations across modalities from mixed visual and language inputs~\citep{Li2019VisualBERT,lxmert,Su2020VLBERT,Chen2019UNITER,Li2020OscarOA}. Among these, \citet{radford2021clip} introduced CLIP, a contrastive language-image pre-training framework that employs language as supervision, demonstrating potential for multi-modal tasks and inspiring subsequent variants for improvement~\citep{Jia2021ALIGN,declip,li2023scaling,albef,li2022blip}.
The evolution of large language models like ChatGPT~\citep{chatgpt} has motivated the development of large vision language models (LVLMs), combining powerful vision encoders like CLIP with large language models such as LLaMa~\citep{touvron2023llama} and Vicuna~\citep{vicuna2023}. Achieved through large-scale modality alignment training on image-text pairs~\citep{Alayrac2022FlamingoAV, awadalla2023openflamingo} and supervised fine-tuning on multi-modal instruction tuning datasets~\citep{liu2023llava, li2023m3it}, resulting LVLMs like GPT-4V~\citep{gpt4v} and Qwen-VL~\citep{Qwen-VL} exhibit promising perceptual and cognitive abilities~\citep{yang2023dawn} for engaging user queries. This paper identifies limitations in CLIP and LVLMs for fine-grained image recognition and proposes the CascadeVLM framework to effectively enhance prediction accuracy by harnessing the advantages of both models.

\paragraph{Fine-grained Image Classification}
Fine-grained image recognition, involving categorization into subordinate classes within a broader category, such as cars~\citep{stanfordcars_dataset} and aircraft models~\citep{fgvc_aircraft_dataset}, demands fine-grained feature learning. Previous work explores diverse strategies, including local-global interaction modules with attention mechanisms~\citep{Fu2017LookCT,Zheng2017LearningMC}, end-to-end feature encoding with specialized training objectives~\citep{Dubey2018maxent,Chang2020TheDI}, and the incorporation of external knowledge bases or auxiliary datasets~\citep{Chen2018KnowledgeEmbeddedRL,Xu2018WeblySupervisedFV}. These approaches offer potential enhancements similar to our CLIP model, which we identify as a future exploration for improved performance.

\paragraph{CLIP Enhancements for Fine-grained Image Classification}
Recent studies have enhanced the CLIP primarily via prompt engineering and pre-training techniques. 
In prompt engineering, CoOp~\citep{DBLP:journals/corr/abs-2109-01134} introduces an innovative method by learning context words as continuous vectors. Extending this idea, CoCoOp~\citep{zhou2022conditional} incorporates a lightweight neural network to generate input-specific tokens for images, further improving model performance. POMP~\cite{li2023scaling} proposes pre-training a general soft prompt on the ImageNet-21K dataset for universal visual tasks. 
Besides, \citet{menon2022visual} instead employs GPT-4 to generate better descriptive prompts for classification, enriching the prompting context for CLIP models. 
In the realm of pre-training, MAWS~\citep{singh2023effectiveness} combines Masked Autoencoder (MAE) pre-training with weakly supervised learning, significantly enhancing the learning efficacy. Similarly, FLIP~\citep{li2023scaling} increases prediction accuracy by masking substantial portions of image patches, facilitating the processing of more image-text pairs within the same timeframe and boosting performance across various tasks.
Different from previous studies, our CascadeVLM explores the integration of LVLMs for leveraging their world knowledge to effectively handle similar classes.


\section{Conclusion}
In this paper, we propose CascadeVLM, harnessing the advantages of CLIP and LVLMs for fine-grained image classification. 
By utilizing CLIP for selecting the potential candidate class, LVLM can make more accurate predictions for image classes with subtle differences.
Experimental results on four benchmarks demonstrate the effectiveness of our proposed framework. Further extension to the few-shot setups showcases the great potential of the cascading framework to leverage the in-context learning ability of LVLMs.

\section*{Limitations}

The efficacy of our CascadeVLM framework hinges critically on the symbiotic interplay between the CLIP model and LVLMs. A key limitation emerges when CLIP's top-K accuracy is insufficient, failing to encompass correct options in LVLM's narrowed candidate set, thereby limiting the scope for enhanced accuracy. Moreover, if CLIP outperforms the LVLM in fine-grained classification, incorporating an LVLM with relatively inferior capabilities may inadvertently diminish overall accuracy. These dynamics underscore the imperative for meticulous selection and alignment of models, ensuring each component's strengths are effectively leveraged within the cascade architecture.

The CascadeVLM framework mainly utilizes the LVLM's extensive familiarity and common knowledge with the dataset. If lack of such knowledge, the accuracy would not be good enough. But we believe the integration of Agent or RAG approches~\cite{DBLP:journals/corr/abs-2005-11401} may offer a solution to this limitation.
Besides, augmenting the context for LVLM with additional information may also be a fact to be explored. Besides the few-shot learning we applied in experiments, we also tested sending the CLIP's prediction scores to LVLM. Unfortunately, this led to a reduction in accuracy, with a nearly 4\% increase in erroneous predictions, attributed to LVLM's over-reliance on CLIP's scores. This highlights the need for a balanced approach to information feeding within the cascade framework.



\bibliography{anthology,custom}

\appendix 

\section*{Appendix}

\section{Prompt Tuning of Qwen}

\subsection{Zero-shot Prompt Tunning of Qwen}

In our experiments, we experimented with various prompt designs to optimize the performance of Qwen in selecting the top-$k$ categories. Two representative prompt styles were identified, each with distinct characteristics and performance implications.

The first prompt style, while intuitive, occasionally led to non-compliant responses where Qwen would select a flower name not listed in the given options, or use an alias instead of the specified name. This approach yielded suboptimal results.

Subsequently, we adapted our prompts to align more closely with the training data of Qwen, where the use of the keyword "options" was prevalent. This adaptation significantly improved compliance and accuracy in the model's responses. Thus for the overall experiment, we use 'PROMPT2'. And for GPT-4V, we applied a similar prompt style but followed the API requirement.

PROMPT 1:
\begin{lstlisting}
Picture 1: <img>....jpg</img>
Please examine the flower image and identify the most suitable flower name corresponding to the image content from the list of flower names below. Remember select only one flower name from the list, and response with the flower name ONLY. Available flower names: [...]
\end{lstlisting}

PROMPT 2:
\begin{lstlisting}
Picture 1: <img>...jpg</img>
Question: What is the flower name? Remember select only one flower name from the options and response with the flower name only. Options: [...]
\end{lstlisting}

\subsection{Few-Shot Prompt Tunning of Qwen}
In the domain of few-shot learning, we conducted experiments with Qwen-VL and observed challenges in its ability to effectively utilize in-context demonstrations and follow instructions. Our experimentation involved different prompt structures in the context of the CLIP-ViT B/32 model with a top-$k = 10$ setting on the Flower102 dataset. 

The initial two prompts led to moderate success, achieving an accuracy of approximately 50\%. However, the implementation of the final prompt design demonstrated a notable improvement, yielding an accuracy close to 68\%. This highlights the impact of prompt design on the model's ability to leverage few-shot learning effectively.

To corroborate the versatility of our CascadeVLM framework, we conducted few-shot learning experiments with GPT-4V. These trials demonstrated the framework's adaptability across different LVLMs, reinforcing its effectiveness in diverse data-rich scenarios.

PROMPT 1:
\begin{lstlisting}
<img>...jpg</img> Question: What is the flower name? Options: [...] Answer: ...
<img>...jpg</img> Question: What is the flower name? Options: [...] Answer: ...
<img>...jpg</img> Question: What is the flower name? Options: [...] Answer: ...
...
<img>...jpg</img> Question: What is the flower name? Answer: ...
\end{lstlisting}

PROMPT 2:
\begin{lstlisting}
Picture 1: <img>...jpg</img> Question: What is the flower name? Options: [...] Answer: ...
Picture 2: <img>...jpg</img> Question: What is the flower name? Options: [...] Answer: ...
Picture 3: <img>...jpg</img> Question: What is the flower name? Options: [...] Answer: ...
...
Picture 4: <img>...jpg</img> Question: What is the flower name? Options: [...] Answer:
\end{lstlisting}

PROMPT 3:
\begin{lstlisting}
Picture 1: <img>...jpg</img> Question: What is the flower name? Answer: ...
Picture 2: <img>...jpg</img> Question: What is the flower name? Answer: ...
Picture 3: <img>...jpg</img> Question: What is the flower name? Answer: ...
...
Picture 4: <img>...jpg</img> Question: What is the flower name? Options: [...] Answer:
\end{lstlisting}

PROMPT 4:
\begin{lstlisting}
Picture 1: <img>...jpg</img> Answer: ...
Picture 2: <img>...jpg</img> Answer: ...
Picture 3: <img>...jpg</img> Answer: ...
...
Picture 4: <img>...jpg</img> Question: What is the flower name? Options: [...] Answer:
\end{lstlisting}

\section{Analysis}
\label{sec:appendix_analysis}

\subsection{Influence of candidate classes number $k$}
\label{subsec:topk_analysis}
In the analysis of the influence of the number of candidate classes, $k$, on classification performance, two distinct configurations of the CLIP model, namely CLIP-ViT-B/16 and CLIP-ViT-L/14, as well as the integration of CLIP ViT-B/32 with Qwen in a cascade framework, have been explored. The investigation reveals dataset-specific optimal settings for $k$. Specifically, for the StanfordCars and FGVC Aircraft datasets, peak performance is observed at a top-10 setting across different configurations, with an interesting shift to top-5 for the FGVC Aircraft dataset when using the ViT-L/14 model, highlighting an enhancement in baseline performance. In contrast, the Flower102 and BirdSnap datasets exhibit optimal results at a top-3 setting, with the Flower102 dataset showing a superior accuracy with the CLIP-ViT-L/14 model, attributed to its intrinsic fine-grained image classification capability. This suggests that the CLIP-ViT-L/14 model's performance surpasses that of the Qwen LVLM in specific cases. Furthermore, the validation performance across different datasets demonstrates a dependency on the chosen value of $k$, indicating a nuanced behavior where the intrinsic properties of each dataset may favor a different range of candidate classes. This behavior underscores the importance of tailoring the cascade framework's parameters to the specific dataset at hand to achieve optimal performance, as evidenced by the gradual improvement in accuracy with a narrower focus in candidate classes for certain datasets, and a discernible peak before a decline in others, suggesting a balance between too few and too many options is crucial for maximizing classification accuracy.

\begin{figure}[t!]
\centering
\includegraphics[width=\linewidth]{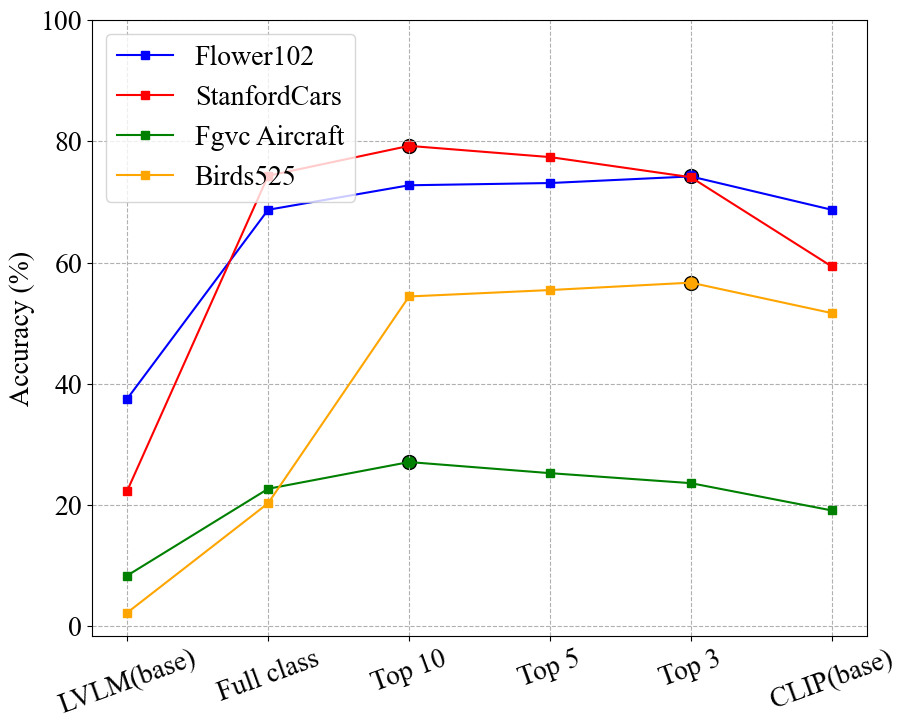}
\caption{Performance changes with varied $k$ with CLIP-ViT-B/32.}
\label{fig:clipvitb32_trending}
\end{figure}

\begin{figure}[t!]
\centering
\includegraphics[width=\linewidth]{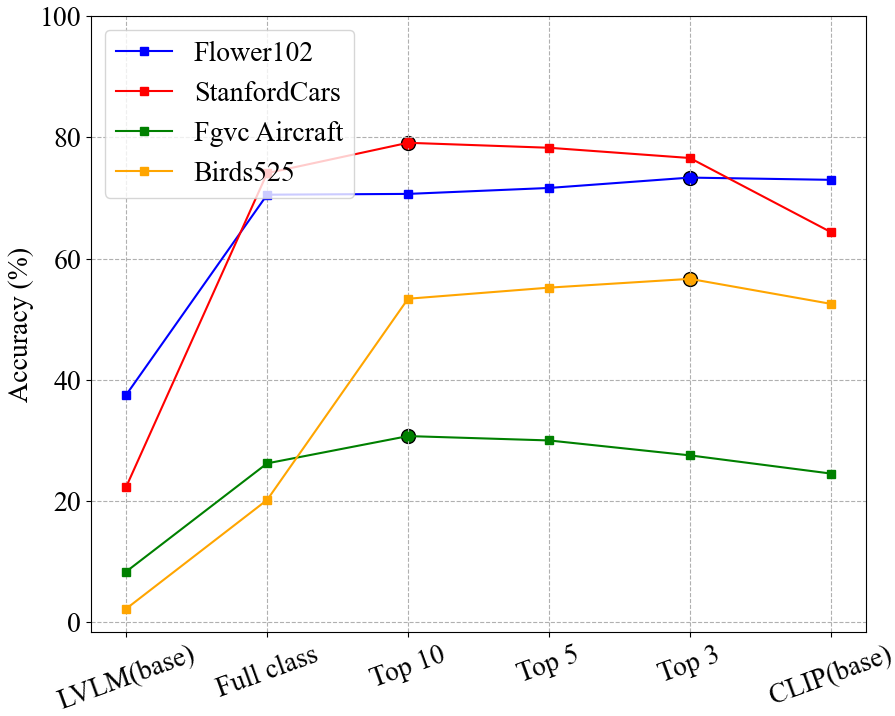}
\caption{Performance changes with varied $k$ with CLIP-ViT-B/16.}
\label{fig:clipvitb16_trending}
\end{figure}

\begin{figure}[t!]
\centering
\includegraphics[width=\linewidth]{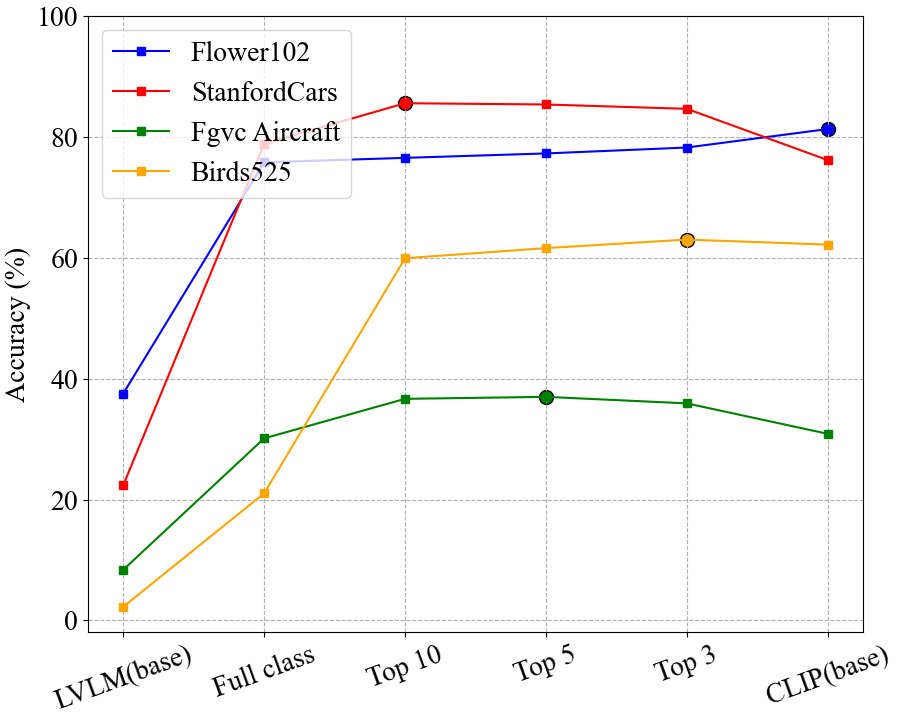}
\caption{Performance changes with varied $k$ with CLIP-ViT-L/14.}
\label{fig:clipvitbl14_trending}
\end{figure}

\subsection{Error Analysis}
\label{subsec:error_analysis}

\begin{figure}[t!]
\centering
\includegraphics[width=\linewidth]{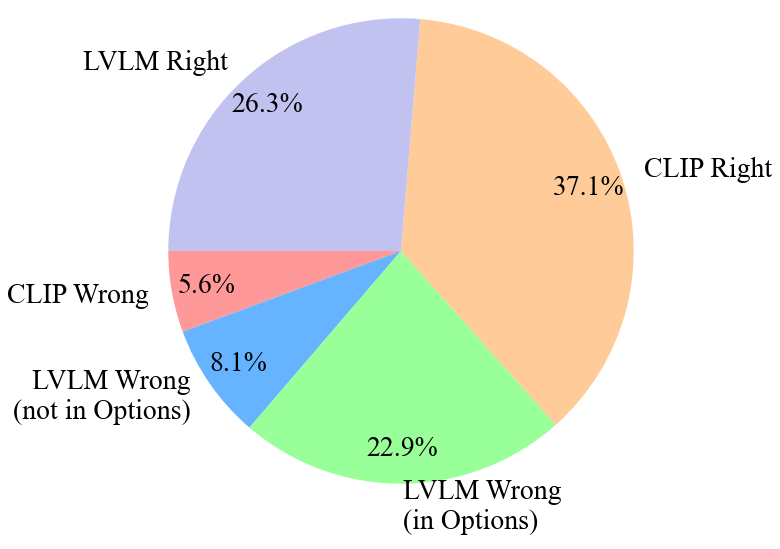}
\caption{Error analysis of the BirdSnap dataset with an entropy threshold of 1.25 and top-k=10. The analysis reveals that despite CLIP including correct options, LVLM frequently misclassifies.}
\label{fig:error_analysis}
\end{figure}

An error analysis was conducted on the BirdSnap dataset using the cascade framework, which incorporates CLIP (ViT-L/14) for initial classification and Qwen as the LVLM for refined categorization with $k=10$, as shown in Figure~\ref{fig:error_analysis}. When entropy is lower than the threshold, prediction is only processed by CLIP, in this case, $148$ misclassifications were noted (\textit{CLIP WRONG}). Otherwise, after the CLIP narrows down the options of classes, the LVLM Qwen would do the final classification. In this case, LVLM resulted in $812$ misclassifications (\textit{LVLM Wrong}), which further breaks down into two categories: $212$ instances where the correct option was not present in the top-10 candidates given by CLIP(\textit{LVLM Wrong $not\ in\ Options$}), and $600$ instances where the correct option was present, but the LVLM failed to identify it (\textit{LVLM Wrong $in\ Options$}).

\end{document}